\documentclass{article}

\usepackage[preprint,nonatbib]{neurips_2020}

\usepackage[utf8]{inputenc} 
\usepackage[T1]{fontenc}    
\usepackage{url}            
\usepackage{booktabs}       
\usepackage{amsfonts}       
\usepackage{nicefrac}       
\usepackage{microtype}      
\usepackage{subfig}
\usepackage{mathtools}
\usepackage{amsmath}
\usepackage[inline]{enumitem}
\usepackage{comment}
\usepackage{amsthm}
\usepackage{amssymb}
\usepackage{tikz}
\usepackage{pgfplots}
\usepackage{algorithm}
\usepackage[noend]{algpseudocode}
\newtheorem{theorem}{Theorem}

\newtheorem{problem}{Problem}
\usepackage{dsfont}
\usepackage{booktabs}
\usepackage{multirow}
\usepackage{caption}
\usepackage{wrapfig,lipsum}
\usepackage{pifont}
\usepackage{todonotes}
\usepackage[hidelinks]{hyperref} 

\usetikzlibrary{automata, arrows,positioning}
\tikzset{every edge/.append style={shorten >= 1pt}}
\tikzset{auto, >= stealth}

\pgfplotsset{compat=1.14}

\DeclareMathOperator{\X}{\mathsf{X}}
\DeclareMathOperator{\U}{\mathsf{U}}
\DeclareMathOperator{\F}{\mathsf{F}}
\DeclareMathOperator{\G}{\mathsf{G}}

\newcommand{\framework}{\textsf{LEXR}}

\usepackage{grffile}

\title{A Formal Language Approach to Explaining RNNs\thanks{The source code is available at \url{https://github.com/bishwamittra/LEXR}}}

\author{%
	Bishwamittra Ghosh\thanks{Work performed at Max Planck Institute for Software Systems, Kaiserslautern, Germany.} \\
	National University of Singapore\\
	Singapore\\
	\texttt{bghosh@u.nus.edu} \\
	 \And
	 Daniel Neider\\
	 Max Planck Institute for Software Systems \\
	 Kaiserslautern, Germany \\
	 \texttt{neider@mpi-sws.org} \\
}

\begin{document}

\maketitle

\begin{abstract}
This paper presents \framework, a framework for explaining the decision making of recurrent neural networks (RNNs) using a formal description language called Linear Temporal Logic (LTL).
LTL is the de~facto standard for the specification of temporal properties in the context of formal verification and  features many desirable properties that make the generated explanations easy for humans to interpret: it is a descriptive language, it has a variable-free syntax, and it can easily be translated into plain English.
To generate explanations, \framework\ follows the principle of counterexample-guided inductive synthesis and combines Valiant's probably approximately correct learning (PAC)  with constraint solving.
We prove that \framework's explanations satisfy the PAC guarantee (provided the RNN can be described by LTL) and show empirically that these explanations are more accurate and easier-to-understand than the ones generated by recent algorithms that extract deterministic finite automata from RNNs. 

\end{abstract}

\section{Introduction}

Recent advances in artificial intelligence and machine learning, especially in deep neural networks, have shown the great potential of algorithmic decision making in a host of different applications.
The inherent black-box nature of today's complex machine learning models, however, has raised concerns regarding their safety, reliability, and fairness.
In fact, the lack of explanations as to why a learning-based system has made a certain decision has not only been identified as a major problem by the scientific community but also by society at large.
A prominent example for this is the European Union, who considers imposing a ``right to explanation'' to future algorithmic decision making~\cite{goodman2017european}.

The task of explaining the decisions of feed-forward networks has received significant attention in the last years.
A recent survey by Du, Liu, and Hu~\cite{du2019techniques} categorizes existing techniques to increase the interpretability of machine learning along two dimensions:
\begin{enumerate*}[label={(\alph*)}]
	\item whether the techniques are intrinsic~\cite{GMM20,ignatiev2018sat,LRMM2015} (i.e., have explainability build into the model) or post-hoc~\cite{lakkaraju2016interpretable,lundberg2017unified,ribeiro2016should} (i.e., explain an existing black-box model) and
	\item whether the explanation is local~\cite{cruz2013deep,olah2018building} (i.e., specific to an input or a restricted region of the input space) or global~\cite{freitas2014comprehensible} (i.e., explaining the behavior on all possible inputs).
\end{enumerate*}
However, comparatively less attention has been put on explaining recurrent neural networks (RNNs), albeit them being a class of neural networks that is routinely used in many practical applications, such as time series prediction~\cite{hewamalage2019recurrent,schmidhuber2005evolino},  speech recognition~\cite{graves2013speech}, and robot control~\cite{mayer2008system}. 

In this work, we address the problem of generating post-hoc explanations for recurrent neural networks (i.e., we assume the RNN to be given as a black-box).
As this problem is extremely challenging in its generality, we focus on so-called \emph{RNN acceptors}~\cite{mayr2018regular,DBLP:conf/icml/WeissGY18}, which are recurrent neural networks whose output is fed into a binary classifier (e.g., a linear classifier, a deep neural network, etc.) and, hence, either ``accept'' or ``reject'' their input.
This kind of neural network has close connections to formal languages, and various approaches to extract explanations in the form of deterministic finite automata (DFAs) have been proposed recently~\cite{mayr2018regular,DBLP:conf/icml/WeissGY18}.
Due to their operational nature, however, DFAs are not always easy to understand---they describe \emph{how} a language is computed but not \emph{what} this language is.
Moreover, the aforementioned DFA extraction techniques tend to produce large automata, containing hundreds of states and transitions, because they search for global explanations of RNN acceptors.

To alleviate these practical shortcomings, we propose a method to generate declarative, local explanations in a formal description language called \emph{Linear Temporal Logic (LTL)}~\cite{pnueli1977temporal}.
Roughly speaking, LTL is an extension of propositional logic that uses temporal modalities, such as $\X$~(``next''), $\F$~(``finally''), $\G$~(``globally''), and $\U$~(``until''), to express properties about sequences in an intuitive and easy-to-understand way.
For instance, the LTL formula $ \F(a \wedge \X (\G b)) $ states that there exists a position in the sequence ($\F$) where proposition $a$ holds and from the next position onwards ($\X$), proposition $b$ always holds ($\G$).
As can be seen from this example, LTL features a variable-free syntax and is very close to plain English, two properties that have contributed greatly to its adoption as the de~facto standard for specifying reactive systems (e.g., see various textbooks on formal verification~\cite{DBLP:books/daglib/0020348,DBLP:reference/mc/2018}).
In addition, Camacho and McIlraith~\cite{DBLP:conf/aips/CamachoM19} have recently proposed to use LTL for AI-related tasks, including plan intent recognition, knowledge extraction, and reward function learning.

\textbf{The main contribution of this paper is a novel  framework, named \framework\ (\underline{L}TL \underline{ex}plains \underline{R}NNs) for explaining the decision making of RNN acceptors using LTL.}
In contrast to the existing DFA extraction methods, \framework\ allows the user to \emph{query} regions of the input space (expressed in LTL) and generates local explanations for the behavior of the RNN acceptor inside such a region.
At its heart lies an iterative procedure that combines inductive methods from the area of machine learning and deductive constraint satisfaction techniques to infer an LTL formula that is a probably approximately correct (PAC)~\cite{DBLP:journals/cacm/Valiant84} local explanation.
Moreover, since the general trend in the literature is to consider smaller formulas to be more interpretable than larger ones~\cite{DBLP:conf/aips/CamachoM19,GMM20,GM19,neider2018learning,DBLP:journals/corr/abs-2002-03668}, \framework\ spends additional computational effort to synthesize formulas of minimal size.

To assess the explanatory performance of \framework, we have implemented a Python prototype and compared it to the DFA extraction method of Mayr and Yovine~\cite{mayr2018regular}, which also uses PAC learning.
Our experimental results on synthetic and  real-world benchmarks show that \framework's explanations are often small and typically more accurate than the DFAs generated by Mayr and Yovine's approach.


\paragraph*{Related Work}
\label{sec:related-works}

In general, methods to increase the interpretability of machine learning can broadly be  categorized along two orthogonal dimensions: intrinsic~\cite{GMM20,ignatiev2018sat,LRMM2015} or post-hoc~\cite{lakkaraju2016interpretable,lundberg2017unified,ribeiro2016should} as well as local~\cite{cruz2013deep,olah2018building} or global~\cite{freitas2014comprehensible}.
The approach proposed here generates model-agnostic (post-hoc) and local explanations.
However, it is slightly different from most other methods for local explanations in that it does not seek explain the decision making relative to (the surrounding of) a particular input but rather inside a user-defined region of interest.
This is a deliberate design choice as the surrounding of a sequence is usually specific to the application domain and should not be defined generically.

Various methods have been proposed to explain the decision making of RNNs and RNN acceptors~\cite{jacobsson2005rule,mayr2018regular,okudono2019weighted,omlin1996extraction,DBLP:conf/icml/WeissGY18}.
Most relevant to our work are the approaches by Weiss, Goldberg, and Yahab~\cite{DBLP:conf/icml/WeissGY18} as well as Mayr and Yovine~\cite{mayr2018regular}, which both use Angluin's L\textsuperscript{$\ast$} algorithm to learn a global explanation in the form of a DFA.
Though these two methods are similar, they differ in the way conformance of the DFA and the RNN acceptor is established: Weiss, Goldberg, and Yahab~\cite{DBLP:conf/icml/WeissGY18} use a partition refinement approach, while Mayr and Yovine~\cite{mayr2018regular} use one inspired by Valiant's probably approximately correct (PAC) learning~\cite{DBLP:journals/cacm/Valiant84}.
We here follow Mayr and Yovine's approach and generate an LTL formula that might make a small number of mistakes but is structurally simpler than an exact explanation.

\framework\ uses a learning algorithm for LTL formulas proposed by Neider and Gavran~\cite{neider2018learning}.
Due to the modular nature of our framework, however, it is easy to substitute this learning algorithm with similar LTL learning algorithms, such as the one by Camacho and McIlraith~\cite{DBLP:conf/aips/CamachoM19}, or even with learning algorithms for more expressive languages, such as the Property Specification Language~\cite{DBLP:journals/corr/abs-2002-03668}.


\section{Preliminaries}
\label{sec:preliminaries}


\paragraph{Recurrent Neural Networks Acceptors}
\label{sec:rnn_preliminaries}

Intuitively, a recurrent neural networks acceptor, \emph{RNN acceptor} for short, is a recurrent neural network whose output is fed into a binary classifier (linear classifier, deep neural networks etc.). 
To make this notion precise, we first introduce the concepts of alphabets and words, which we borrow from the theory of formal languages.
An \emph{alphabet} simply is a finite set $\Sigma \subset \mathbb R^{d_i}$, where $d_i \in \mathbb N \setminus \{ 0 \}$ is the input dimension of the RNN (i.e., the number of its input neurons).
We call each element of $\Sigma$ a \emph{symbol} and encourage the reader to think of each symbol as a one-hot encoded vector.
A \emph{word} is a finite sequence $u = a_1 \ldots a_n$ of symbols (i.e., $a_i \in \Sigma$ for each $i \in \{ 1, \ldots, n \}$).
Moreover, we denote the empty word (i.e., the empty input sequence) by $\lambda$ and the set of all finite words over the alphabet $\Sigma$ by $\Sigma^\ast$.
A subset $L \subseteq \Sigma^\ast$ is called a \emph{language}, and $L_1 \oplus L_2 = L_1 \setminus L_2 \cup L_2 \setminus L_1$ denotes the \emph{symmetric difference} of the two languages $L_1, L_2 \subseteq \Sigma^\ast$.

Since we are interested in providing explanations for the input-output behavior of an RNN acceptor (but not how they work internally), we define RNN acceptors in an abstract form, which follows the work of Mayr and Yovine~\cite{mayr2018regular} as well as Weiss, Goldberg, and Yahav~\cite{DBLP:conf/icml/WeissGY18}.
More precisely, we view an RNN $R$ as a function $g_R \colon \mathbb R^{d_s} \times \mathbb R^{d_i} \to \mathbb R^{d_s}$, which takes an internal state $s \in \mathbb R^{d_s}$ of the RNN and  a symbol $a \in \Sigma$ as inputs and produces a new state $s' \in R^{d_s}$ as output.
Similarly, we view the binary classifier $C$ as a function $f_C \colon \mathbb R^{d_s} \to \{ 0, 1 \}$ that takes an internal state $s \in \mathbb R^{d_s}$ of the RNN as input and maps it to either class $0$ (``accept'') or class $1$ (``reject'').

An \emph{RNN acceptor} is now a pair $\mathcal R = (R, C)$ consisting of a RNN $R$ and a binary classifier $C$.
To define the output of such an RNN acceptor, we first extend the function $g_R$ to words: $g_R^\ast(s, \lambda) = s$ and $g_R^\ast(s, ua) = g_R(g_R^\ast(s, u), a)$ for all states $s \in \mathbb R^{d_s}$, words $u \in \Sigma^\ast$, and symbols $a \in \Sigma$.
This allows us to define the \emph{language of an RNN acceptor $\mathcal R$} as the set
\[ L(\mathcal R) = \{ u \in \Sigma^\ast \mid f_C(g_R^\ast(s_I, u))= 1\} \subseteq \Sigma^\ast,  \]
where $s_I \in \mathbb R^{d_s}$ is a fixed initial state of the RNN (typically the null vector). 
In other words, the language of $\mathcal R$ contains all words (input sequences) for which the RNN acceptor outputs class $1$.

The task we want to solve in this paper is to find a human-interpretable description of $L(\mathcal R)$.
To this end, we use the formal description language Linear Temporal Logic, which is introduced next.

\paragraph{Linear Temporal Logic}
\label{sec:LTL}

The description language \emph{LTL}, short for \emph{Linear Temporal Logic}~\cite{pnueli1977temporal}, is an extension of propositional logic that allows reasoning about sequences.
To this end, LTL introduces temporal modalities, such as $\X$ (``next''), $\F$ (``finally''), $\G$ (``globally''), and $\U$ (``until'').
This allows expressing properties of sequences in a very intuitive and easy-to-understand way.
For instance, the property ``the sequence contains the symbol $a$ at some point'' is expressed by $\F a$, while the property ``every $a$ is immediately followed by the symbol $b$'' is expressed by $\G (a \rightarrow \X b)$.

Formally, formulas in LTL are defined according to the following grammar:
\[ \varphi \Coloneqq a \in \Sigma \mid \lnot \varphi \mid \varphi \lor \varphi \mid \X \varphi \mid \varphi \U \varphi. \] 
We also allow syntactic sugar in form of the formulas $\top$ (``true''), $\bot$ (``false''), $\varphi \land \psi$, $\varphi \rightarrow \psi$, which are defined in the usual way.
Moreover, we allow the formulas $\F \varphi \coloneqq \top \U \varphi$ and $\G \varphi \coloneqq \lnot \F \lnot \varphi$.

Although LTL formulas are originally interpreted over infinite sequences, we use a semantics over finite words, introduced by Giacomo and Vardi~\cite{DBLP:conf/ijcai/GiacomoV15}.
More precisely, given a word $u = a_1 \ldots a_n$ and a position $i \in \{ 0, \ldots, n \}$, we define a relation $\models$ that formalizes when the suffix of $u$ starting at position $i$ \emph{satisfies} an LTL formula:
$(u, i) \models a$ iff $a_i = a$; $(u, i) \models \lnot \varphi$ iff $(u, i) \not\models \varphi $; $(u, i) \models \varphi_1 \lor \varphi_2$ iff $(u, i) \models \varphi_1$ or $(u, i) \models \varphi_2$; $(u, i) \models \X \varphi$ iff $i < n$ and $(u, i+1) \models \varphi$; and $(u, i) \models \varphi_1 \U \varphi_2$ iff there exists a $j \in \{ i, \ldots, n \}$ such that $(u, j) \models \varphi_2$ and $(u, k) \models \varphi_1$ for each $k \in \{ i, \ldots, j-1 \}$.
We say that an LTL formula $\varphi$ \emph{satisfies} a word $u \in \Sigma^\ast$ if $(u, 1) \models \varphi$ (i.e., $u$ satisfies $\varphi$ starting at the first position); we then write $u \models \varphi$.
Similar to an RNN acceptor, an LTL formula $\varphi$ defines a language $L(\varphi) = \{ u \in \Sigma^\ast \mid u \models \varphi\} \subseteq \Sigma^\ast$, which consists of all words $u \in \Sigma^\ast$ that satisfy $\varphi$.

We define the \emph{size} of an LTL formula $\varphi$, denoted by $|\varphi|$, as the number of its unique sub-formulas (e.g., the formula $a \lor \X a$ is of size three as it has three unique sub-formulas: $a$, $\X a$, and $a \lor \X a$).
In the remainder, we use the size of an LTL formula as a metric of how easy it is for humans to comprehend a formula: the smaller a formula, the easier it is to comprehend.
Although the definition of interpretability of formulas is not unanimously agreed upon, it is a general trend in the literature to consider smaller formulas to be more interpretable than larger ones (we refer the reader to recent work on learning interpretable formulas for a more in-depth discussion~\cite{DBLP:conf/aips/CamachoM19,GMM20,GM19,neider2018learning,DBLP:journals/corr/abs-2002-03668}).

%

Finally, it is important to point out that the expressive power of LTL is that of first-order logic over words~\cite{DBLP:journals/corr/Rabinovich14}.
Although this is a proper subset of regular languages, 
LTL is powerful enough to express most real-world properties and has become the de~facto standard for specifying properties of safety-critical systems~\cite{DBLP:books/daglib/0020348}.
Note, however, that LTL is in general not powerful enough to express the language of an RNN acceptor because feed-forward networks (with a non-polynomial activation function) can approximate any function~\cite{DBLP:journals/nn/LeshnoLPS93}.
Section~\ref{sec:theoretical-analysis} discusses how \framework\ handles such situations.


\section{Problem Statement}
\label{sec:problem-statement}

Given an RNN acceptor $\mathcal R$, we would ideally like to construct an explanation of $\mathcal R$ in the form of an LTL formula $\varphi$ satisfying $L(\varphi) = L(\mathcal R)$.
However, this is challenging for two reasons:
\begin{enumerate*}[label={(\arabic*)}]
	\item an RNN acceptor can have a drastically different behavior in various parts of its input space and, consequently, an LTL formula explaining this behavior on all possible input-words has to be very complex (i.e., large) as well;
	\item even in restricted parts of the input space, an RNN acceptor can have a very complex behavior and every LTL formula that captures this behavior precisely is necessarily large (if one exists at all).
\end{enumerate*}
Towards the goal of understanding an RNN  acceptor, however, large LTL formulas are of little help because they are arguably as hard to understand as the RNN acceptor itself.

To address the first challenge, we follow a common approach in the literature and seek local explanations for the behavior of an RNN acceptor rather than a global one~\cite{freitas2014comprehensible,mayr2018regular,DBLP:conf/icml/WeissGY18}.
What we mean by this is to construct an LTL formula that explains the language of an RNN acceptor in a particular region of its input space, which we call a \emph{query}.
To specify such a query, we again use LTL.
Thus, our problem becomes the following: given an RNN acceptor $\mathcal R$ and a query $\psi$, construct an LTL formula $\varphi$ such that $L(\varphi) = L(\mathcal R) \cap L(\psi)$---in other words, the formula $\varphi$ should describe the exact language of the RNN acceptor $\mathcal R$ inside the query $\psi$.
For instance, given an RNN acceptor $\mathcal R$ with $L(\mathcal R) = \{ a, b, aa, bb, aaa, bbb, \ldots \}$ and the query $\psi \coloneqq \F a$, which asks for the behavior of $\mathcal R$ on words that contain the symbol $a$ at least once, we expect the LTL formula $\varphi \coloneqq \G a$ as an answer (note that $L(\varphi) = \{ a, aa, aaa, \ldots \}$).
Note that a word in the symmetric difference $L(\varphi) \oplus (L(\mathcal R) \cap L(\psi))$ is one on which the explanation $\varphi$ makes a mistake.

We address the second challenge by designing an algorithm that follows Valiant's popular PAC framework~\cite{DBLP:journals/cacm/Valiant84} and produces \emph{probably approximately correct (PAC)} explanations rather than exact ones.
To make this idea precise, let $\mathcal D$ be an arbitrary probability distribution over the set $\Sigma^\ast$ of all finite words.
Then, we define the \emph{explanation error} of an LTL formula $\varphi$ with respect to an RNN acceptor $\mathcal R$ and a query $\psi$ to be the probability of a word $u \in \Sigma^\ast$ belonging to the symmetric difference $L(\varphi) \oplus (L(\mathcal R) \cap L(\psi))$, denoted by $\mathbf{P}_\mathcal D \bigl( \varphi \oplus \mathcal R \cap \psi \bigr)$. 
Moreover, given an \emph{approximation parameter} $\varepsilon \in (0, 1)$, we say that an LTL formula $\varphi$ is an \emph{$\varepsilon$-explanation} for $\mathcal R$ and $\psi$ if $\mathbf{P}_\mathcal D \bigl( \varphi \oplus \mathcal R \cap \psi \bigr) < \varepsilon$.
Note that $\varepsilon$-explanations can be much more succinct than exact ones since they are allowed to make a small number of errors in favor of a simpler explanation.
This makes approximate explanations particularly well-suited for our purpose.

Our goal in this paper is now to design an algorithm that produces an $\varepsilon$-explanation with a sufficiently high confidence, as stated next.

\begin{problem} \label{prob:PAC-explanations}
Given a probability distribution $\mathcal D$ over $\Sigma$, an RNN acceptor $\mathcal R$, a query $\psi$, an approximation parameter $\varepsilon \in (0, 1)$, and confidence parameter $\delta \in (0, 1)$, compute an LTL formula $\varphi$ such that $\varphi$ is an $\varepsilon$-explanation for $\mathcal R$ and $\psi$ with probability at least $1-\delta$.
\end{problem}

In the next section, we design an algorithm for solving Problem~\ref{prob:PAC-explanations}.
Since this algorithm works with any distribution, we omit the subscript $\mathcal D$ in the remainder of this paper.



\section{{\framework}: A Framework for Explaining RNN Acceptors Using LTL}
\label{sec:framework}

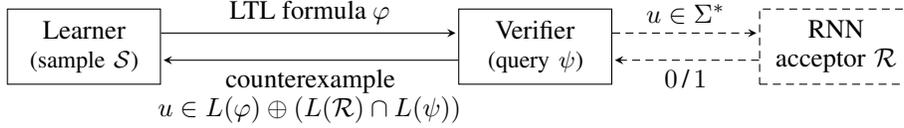
\begin{figure}[t]
	\centering
	\begin{tikzpicture}
		\node[draw, text width=18mm, minimum height=10mm, align=center] (learner) at (0, 0) {Learner \\ \footnotesize (sample $\mathcal S$)};
	
		\node[draw, text width=18mm, minimum height=10mm, align=center] (verifier) at (6, 0) {Verifier \\ \footnotesize (query $\psi$)};
	
		\node[draw, text width=18mm, minimum height=10mm, align=center, densely dashed] (RNN) at (10, 0) {RNN acceptor $\mathcal R$};
	
		\path[->] (learner.10) edge node {LTL formula $\varphi$} (verifier.170);
		\path[->] (verifier.190) edge node[align=center] {counterexample \\ $u \in L(\varphi) \oplus (L(\mathcal R) \cap L(\psi))$} (learner.350);
		\path[->, densely dashed] (verifier.10) edge node {$u \in \Sigma^\ast$} (RNN.170);
		\path[->, densely dashed] (RNN.190) edge node {$0$\,/\,$1$} (verifier.350);
	\end{tikzpicture}
	\caption{Feedback loop of \framework.} \label{fig:feedback-loop}
	\vspace{-15 pt}
\end{figure}

In this section, we discuss the main contribution of this work, \framework, a framework for explaining RNN acceptors using the formal description language LTL. 
As shown in Figure~\ref{fig:feedback-loop}, \framework\ follows the principle of \emph{iterative passive learning}~\cite{DBLP:journals/tc/BiermannF72}, which is also called \emph{counterexample-guided inductive synthesis}~\cite{CEGIS}.
At its core is a feedback loop between two entities: a \emph{learner}, who is agnostic to the RNN acceptor and the query, and a \emph{verifier}\footnote{In the jargon of Angluin~\cite{angluin1987learning}, our verifier is a teacher that answers equivalence queries in an approximate manner using a set of randomly generated membership queries.}, who can sample the RNN acceptor (inside and outside the query).
In every iteration of the loop, the learner conjectures an LTL formula that it infers from the data it has gathered so far.
The verifier, on the other hand, checks whether the proposed LTL formula is an $\varepsilon$-explanation using a random sampling technique inspired by PAC learning~\cite{DBLP:journals/cacm/Valiant84} and recently used to extract finite automata from RNN acceptors~\cite{mayr2018regular}.
If the proposed LTL formula and the RNN acceptor conform on all of these random samples, then the feedback loop stops and \framework\ returns the current LTL formula; as we show later, this formula is indeed an $\varepsilon$-explanation with probability at least $1 - \delta$.
However, the verifier might detect that the RNN acceptor and the proposed LTL formula have a different behavior on one (or more) random samples (i.e., a random sample $u \in \Sigma^\ast$ belongs to $L(\varphi) \oplus (L(\mathcal R) \cap L(\psi))$).
If this happens, the verifier returns such words as so-called \emph{counterexamples} to the learner, who uses this new information to refine its conjecture in the next iteration.
\framework\ repeats this process until a conjecture passes the verification step (or some user-defined computational budget has been exhausted).

The remainder of this section describes \framework\ in detail.
Sections~\ref{sec:verifier} and \ref{sec:learner} present the verifier and the learner, respectively.
Section~\ref{sec:theoretical-analysis} shows the correctness of our framework and discusses the situation that \framework\ is stopped early.

\subsection{The Verifier}
\label{sec:verifier}

The task of the verifier is to check whether a conjectured LTL formula $\varphi$ is indeed an explanation for the RNN acceptor $\mathcal R$ and the query $\psi$ in the sense that $L(\varphi) = L(\mathcal R) \cap L(\psi)$.
Since this equality is very hard to check in general, our verifier provides an approximate answer based on a finite suite of test inputs, which it draws randomly from $\Sigma^\ast$ according to the underlying probability distribution $\mathcal D$.
The exact size of the test suite depends on the number of incorrect conjectures the verifier has already received and follows the approach used to extract automata from RNN acceptors~\cite{mayr2018regular}.
More precisely, if $\varphi$ is the conjecture of iteration $i \in \mathbb N \setminus \{  0 \}$ (i.e., the verifier has already received $i-1$ incorrect explanations), the verifier generates a test suite $T_i$ of cardinality $r_i = \lceil\frac{1}{\varepsilon}(i \cdot \ln{2} - \ln{\delta})\rceil$.

Once the test suite $T_i$ has been created, the verifier checks whether a word from $T_i$ witnesses a difference between the languages $L(\varphi)$ and $L(\mathcal R) \cap L(\psi)$.
In other words, the verifier checks whether there exists a word $u \in T_i$ such that $u \in L(\varphi) \oplus (L(\mathcal R) \cap L(\psi))$---we call such a word a \emph{counterexample}.
If a counterexample is found, the verifier stops and returns it to the learner (optionally, the teacher can continue to search for additional counterexamples and return them in a bulk).
If no counterexample is found, on the other hand, the overall feedback loop terminates and \framework\ returns the current LTL formula $\varphi$.
As we show in Section~\ref{sec:theoretical-analysis}, this formula is in fact an $\varepsilon$-explanation with probability at least $1 - \delta$.


To decide whether a word $u \in T_i$ is a counterexample, our verifier checks the equivalent condition
\[ u \notin L(\varphi) \quad\text{if and only if}\quad u \in L(\mathcal R) \text{ and } u \in L(\psi). \]
This can be done by means of three efficient membership tests: $u \in L(\mathcal R)$, $u \in L(\varphi)$, and $u \in L(\psi)$.

We can perform the first membership test in a straightforward manner by passing the word $u$ through the RNN acceptor $\mathcal R$ and checking the resulting class label.
If the label is $1$, then $u \in L(\mathcal R)$; if the label is $0$, then $u \notin L(\mathcal R)$.
Note that this procedure can naturally be parallelized on a GPU.

The latter two membership tests can be performed using dynamic programming.
Given a finite word $u$ and an LTL formula $\eta$, the key idea is to generate a two-dimensional table $\tau$ such that each table entry $\tau[i, \chi]$ indicates whether the suffix of $u$ starting at position $i$ satisfies the sub-formula $\chi$ of $\eta$ (the answer to the membership test is then the entry $\tau[1, \eta]$).
The computation of the individual table entries follows the semantics of LTL (a table entry for a complex formula is computed based on table entries for its simpler sub-formulas) and can easily be parallelized.
In fact, the membership test for LTL on finite words has been shown to fall into the complexity class \textsf{NC}~\cite{DBLP:conf/icalp/KuhtzF09}, which is the class of all problems that can be efficiently solved on a parallel computer.

\subsection{The Learner}
\label{sec:learner}

The task of the learner is to produce an LTL formula from the counterexamples  it has gathered so far.
To this end, the learner maintains a finite set $\mathcal S \subset \Sigma^\ast \times \{ 0, 1 \}$, called \emph{sample}, in which it collects all counterexamples together with their correct classification ($0$ or $1$).
The correct classification can be derived in a straightforward manner.
If a word $u$ was returned as a counterexample, then because
\begin{enumerate*}[label={(\alph*)}]
	\item $u \in L(\varphi)$ and $u \notin L(\mathcal R) \cap L(\psi)$ or
	\item $u \notin L(\varphi)$ and $u \in L(\mathcal R) \cap L(\psi)$.
\end{enumerate*}
In the former case, the expected classification is $0$ (since we have to ensure $u \notin L(\varphi')$ for any future conjecture $\varphi'$), while it is $1$ in the latter case.

On a technical level, our learner is a slight modification of a recent algorithm for learning LTL formulas from infinite (i.e., ultimately repeating) words proposed by Neider and Gavran~\cite{neider2018learning}.
The core idea of this---and, hence, our---algorithm is to reduce the learning problem to a series of satisfiability problems in propositional logic and then use a highly-optimized SAT solver to search for a solution. 
Although this approach has the drawback of a high computational complexity, it possesses a crucial feature that is indispensable in our setting: it learns an LTL formula that is \emph{minimal} and, hence, easy for humans to understand.
Moreover, Neider and Gavran have empirically shown that their algorithm can effectively learn minimal LTL formulas in real-world applications. 

Given a sample $\mathcal S$, our algorithm generates a series of propositional formulas $\Phi_n^\mathcal S$, where $n \in \mathbb N \setminus \{ 0 \}$ is a parameter referring to the size of the prospective LTL formula, that have the following two properties:
\begin{enumerate*}[label={(\arabic*)}]
	\item the propositional formula $\Phi_n^\mathcal S$ is satisfiable if and only if there exists an LTL formula of size at most $n$ that is consistent with $\mathcal S$; and
	\item if $\Phi_n^\mathcal S$ is satisfiable, then a satisfying assignment of its variables carries sufficient information to extract a consistent LTL formula of size at most $n$.
\end{enumerate*}
By using a binary search over the parameter $n$, we obtain an effective algorithm for learning a minimal consistent LTL formula from a given sample.
Note that such an LTL formula always exists as LTL is expressive enough to precisely characterize any finite set of words.

The formula $\Phi_n^\mathcal S$ used in our algorithm resembles the one proposed by Neider an Gavran closely.
A major difference is, however, that for LTL on finite words, the next-operator $\X$ has a different semantics at the end of a word (as there is no next position)---a situation that does not occur in the case of infinite words.
To account for this, we have slightly adapted the corresponding constraints in the formula $\Phi_n^\mathcal S$.
We refer the reader to the original algorithm by Neider and Gavran~\cite{neider2018learning} for a complete description of how the formula $\Phi_n^\mathcal S$ is constructed.

\subsection{Theoretical Analysis}
\label{sec:theoretical-analysis}

Clearly, the feedback loop of \framework\ either stops after a finite number of iterations and returns an LTL formula, or it repeats forever (the latter happens if LTL is not expressive enough to explain the language of an RNN acceptor inside a query).
If \framework\ returns a formula $\varphi^\star$, say after $m \geq 1$ iterations, then we claim that $\varphi^\star$ is an $\varepsilon$-explanation with probability at least $1 - \delta$.
To prove that this is in fact true, we observe that the probability of $\varphi^\star$ not being an $\varepsilon$-explanation (i.e., $\mathbf{P}_\mathcal D(\varphi^\star \oplus \mathcal R \cap \psi) \geq \varepsilon$) even if \emph{all} test inputs had passed \emph{all} of the $m$ calls to the verifier is at most
\[ \sum_{i = 1}^m (1 - \varepsilon)^{r_i} \leq \sum_{i = 1}^m e^{-\varepsilon r_i} \leq \sum_{i = 1}^m 2^{-i} \delta \leq \delta. \]
Thus, $\varphi^\star$ is indeed an $\varepsilon$-explanation with probability at least $1 - \delta$, which proves our main result.

\begin{theorem}
Given a probability distribution $\mathcal D$ over $\Sigma$, an RNN acceptor $\mathcal R$, a query $\psi$, an approximation parameter $\varepsilon \in (0, 1)$, and confidence parameter $\delta \in (0, 1)$, \framework\ computes an LTL formula $\varphi$ such that $\varphi$ is an $\varepsilon$-explanation for $\mathcal R$ and $\psi$ with probability at least $1-\delta$.
\end{theorem}

It is worth noting that if \framework\ stops and returns an $\varepsilon$-explanation, the number of test inputs generated by the verifier is bounded by 
$ \sum_{i = 1}^m \lceil r_i \rceil \leq \sum_{i = 1}^m 1 + \frac{1}{\epsilon}(i \cdot \ln{2} - \ln{\delta}) \in \mathcal O ( m + \frac{1}{\varepsilon} [ m^2 + m \cdot \ln{\frac{1}{\delta}} ] )$.
Due to the fact that LTL is less expressive than the class of RNN acceptors, however, it can happen that the language $L(\mathcal R) \cap L(\psi)$ cannot be $\varepsilon$-approximated by an LTL formula, in which case \framework\ might not terminate.
To account for such situations, we stop \framework\ after a user-defined number of iterations (or when a user-defined computational budget has been exhausted) and return the last LTL formula $\varphi$ that the learner has conjectured.

Clearly, we can then no longer hope that $\varphi$ has the desired $(\varepsilon, \delta)$-guarantee, but this explanation still holds statistical meaning.
To make this precise, let us assume that we stopped \framework\ after iteration~$i$ and the test suite $T_i$ contained $k > 0$ counterexamples.
Using the same derivation as in the work of Mayr and Yovine~\cite{mayr2018regular}, we can still accept the hypothesis that $\varphi$ is an $\varepsilon$-explanation with confidence $\delta'> \binom{r_i}{k} e^{-\varepsilon(r_i - k)}$.
Similarly, we can accept the hypothesis that $\varphi$ is an $\varepsilon'$-explanation with probability at least $1 - \delta$ for every $\varepsilon' > \frac{1}{r_i - 1} ( \ln{\binom{r_i}{k}} - \ln{\delta} )$.
Unfortunately, the values for $\varepsilon'$ and $\delta'$ might be larger than $1$ in practice (as observed in our experiments and reported by Mayr and Yovine~\cite{mayr2018regular}).
Therefore, we evaluate the quality of sub-optimal explanations using a different approach, as explained in the next section.


\section{Experiments}
\label{sec:experiments}

In this section, we evaluate the explanatory capabilities of \framework\ and compare it to the current state-of-the-art explanation method of Mayr and Yovine~\cite{mayr2018regular}, which produces explanations in terms of DFAs. 
To this end, we have implemented a Python prototype based on the LTL learning algorithm by Neider and Gavran~\cite{neider2018learning}.
Moreover, we have modified Mayr and Yovine's method such that it can handle queries.
To simplify the implementation of the verifier, we did not use the dynamic programming solution described in Section~\ref{sec:verifier} but a similar technique based on a translation of LTL into DFAs~\cite{francesco_fuggitti_2019_2598764}.
Both implementations use the same verifier (though Mayr and Yovine's method also asks for the classification of individual test inputs).

Since assessing the quality of an explanation is difficult (as we do not know one), we use a statistical test that computes the accuracy of an explanation on a separate test set $T \subset \Sigma^\ast$. 
Slightly deviating from the standard definition, we define accuracy as the fraction of words $u \in T$ that satisfy $u \in L(\varphi)$ if and only if $u \in L(\mathcal R) \cap L(\psi)$ (i.e., $\varphi$ and $\mathcal R$ agree).
To determine whether an LTL formula is easier or harder to interpret as a DFA, we compare their sizes (i.e., number of sub-formulas and states), following the principle ``smaller is easier to interpret`` typically used in the literature~\cite{DBLP:conf/aips/CamachoM19,GMM20,GM19,neider2018learning,DBLP:journals/corr/abs-2002-03668}.

We have conducted experiments with six different RNN acceptors, three RNN acceptors for synthetic languages and three RNN acceptors for real-word problems (valid email addresses~\cite{DBLP:conf/icml/WeissGY18}, an alternating bit protocol~\cite{mayr2018regular}, and words of balanced parentheses~\cite{DBLP:conf/icml/WeissGY18}).
We here report summaries of our experiments on two RNN acceptors for synthetic languages and one for the language of balanced parentheses. Details for all experiments can be found in the Appendix.

We have conducted our experiments on an Intel Xeon E7-8857 v2 CPU with 48 cores and $1.5$\,TB of RAM running on a 64bit Linux distribution based on Debian. 
We have set $\varepsilon$  and $\delta$  to  $0.05$ and the timeout to $400$\,s.
Since the verifier implements a probabilistic algorithm, we have repeated each experiment $ 250 $ times and report an average of all numerical quantities.%
\footnote{
Although we assume the RNN acceptors to be black-boxes, we briefly provide information about their training.
All RNNs were LSTM networks with $3$ layers and $10$ hidden units, while the linear classifier was a linear threshold function with threshold $0.5$.
For each synthetic (real-world) problem, we have generated $6,000$ ($20,000$) random words and split them into training and test sets with an $8:2$ ratio.
We used the Adam optimizer with a learning rate of $0.001$ and a stopping threshold of $0.005$ to train the networks.
On the test set, the RNN acceptors have achieved more than $ 99.9\%$ accuracy.
}

\textbf{Synthetic Benchmarks~~}
We have trained two RNN acceptors for the languages  $\F(a)$ and $\F(a \land \X b)$ over the alphabet $\Sigma = \{a, b, c \}$, respectively.
For each problem, we have considered four simple queries, as shown in Table~\ref{tab:example-synthetic}.
This  table also shows the averaged results of \framework\ as well as one explanation that has an accuracy close to the average accuracy of all $250$ repetitions.
For explanations in form of DFAs, we report the average results as well.

\begin{table}[t]
	\caption{Averaged results on RNN acceptors for synthetic languages. ``---'' indicates a time out.} 	\label{tab:example-synthetic}
	\scriptsize
	\begin{center}
		\begin{tabular}{lllcrrrrrrrr}
			\toprule
			Problem& Query & \multicolumn{4}{c}{LTL} & \multicolumn{3}{c}{DFA} \\
			\cmidrule(r){3-6}
			\cmidrule(r){7-9}
			& & Explanation & $ |\varphi| $ & Acc(\%) &  Time (s) & |Q| & Acc(\%) &  Time (s)\\
			\midrule

			$  \F (a)  $  
			& $ \top $ & $ \F  a $ & $ 2.0 $ & $ 100.0 $ & $ 3.09 $ & $ 2.0 $ & $ 100.0 $ & $ 0.29 $ \\
			& $ \bot $ & $ \bot $ & $ 1.0 $ & $ 100.0 $ & $ 0.86 $ & $ 80.9 $ & $ 44.7 $ & $ \text{\textemdash} $ \\
			& $ \F (b) $ & $ (\F  a)  \wedge  (\F  b) $ & $ 5.0 $ & $ 100.0 $ & $ 13.48 $ & $ 27.3 $ & $ 98.5 $ & $ 218.21 $ \\
			& $ \F ( \neg b) $ & $ \F  a $ & $ 2.0 $ & $ 100.0 $ & $ 3.06 $ & $ 2.0 $ & $ 100.0 $ & $ 0.28 $ \\
				
			\midrule
		
			$  \F (a  \wedge  \X (b))  $ 
			& $ \top $ & $ \F  (a  \wedge  (\X  b)) $ & $ 5.0 $ & $ 100.0 $ & $ 15.98 $ & $ 3.0 $ & $ 100.0 $ & $ 0.39 $ \\
			& $ \bot $ & $ \bot $ & $ 1.0 $ & $ 100.0 $ & $ 0.85 $ & $ 1.0 $ & $ 100.0 $ & $ 0.28 $ \\
			& $ \F (c) $ & $ (\F  (\X  b))  \wedge  (a \U  ( \neg  (\X  b))) $ & $ 6.9 $ & $ 56.9 $ & $ \text{\textemdash} $ & $ 43.7 $ & $ 99.0 $ & $ 381.67 $ \\
			& $ \G (a) $ & $ \bot $ & $ 1.0 $ & $ 100.0 $ & $ 0.84 $ & $ 1.0 $ & $ 100.0 $ & $ 0.28 $ \\
			
			\bottomrule
		\end{tabular}
	\end{center}
\end{table} 

If the query is true ($ \top $), the explanation has to describe the language of the RNN acceptor.
In contrast, when the query is false  ($ \bot $), the explanation should be false as well.
In all synthetic benchmarks, we have found the LTL explanations to be $100\%$ accurate for these two queries.
In fact, we observe that \framework\ produced an explanation with $100\%$ accuracy for all but one query.
The average accuracy, size of explanations, and running times are slightly in favor of \framework. 
However, there is one case where the DFA explanation has a higher average accuracy than the LTL explanation: for the language $ \F(a \wedge \X(b)) $ and the query $ \F(c) $, the DFA has on average $43.7$ states with $ 99\%$ accuracy, whereas the LTL explanation has on average a size of $6.9$ with $56.9\%$ accuracy.

Let us now analyze one query in detail.
For the language $\F a$ and the query $\F(b)$, the learned (and expected) LTL explanation is $\F(a) \wedge \F(b)$.
The average size of this explanation is $5$ and its accuracy is $100\%$.
\framework\ has on average taken around $13.5$\,s for this explanation ($11.1$\,s for the learner and $2.4$\,s for the verifier), and the verifier has generated a total of around $1,400$ random words before it certifies the PAC-guarantee.
In contrast, Mayr and Yovine's method takes on average $218$\,s and generates a DFA with $27.3$ states with $98.5\% $ accuracy.

\textbf{Balanced Parentheses~~}
For this benchmark, the RNN acceptor is trained to predict whether a word has balanced parentheses or not.
We have used the alphabet $\Sigma = \{ l, r, a \}$ where $l$ and $r$ are placeholders for the left parenthesis ``('' and the right parenthesis ``)'', respectively, and $a$ stands for non-parenthesis symbols.
For our evaluation, we have considered twelve queries, which define different valid and invalid properties of balanced parentheses:

\begin{enumerate*}
	\item $  \F (l  \wedge  \X (\G ( \neg r)))  $,
	\item $  \F (l)  \wedge  \F (r)  \wedge  \F ( (l\vee a) \U  r )  $,
	\item $  \F (l)  \wedge  \F (r)  \wedge   \neg (\F ( (l\vee a) \U  r ))  $,
	\item $  \G (a)  $,\\
	\item $  \G (l  \rightarrow \F (r))  $,
	\item $  \G (l  \rightarrow  \neg (\F (a \vee  r)))  $,
	\item $  \G (l)  $,
	\item $  a \U  r  $,
	\item $  \bot  $,
	\item $  r  $,
	\item $  \top  $, and
	\item $   \neg \F (l\vee r)  $.
\end{enumerate*}

\pgfplotsset{
	barplot2/.style={
		xbar=0pt,
		bar width=3pt,
		ytick distance=1,
		tick label style={font=\scriptsize},
		tick align=outside,
		xlabel style={font=\footnotesize\strut, yshift=1mm},
		ylabel style={font=\footnotesize, yshift=-1mm},
		axis x line=bottom,
		axis y line=left,
		axis line style={-},
		xmajorgrids=true,
		major grid style={color=gray!20, very thin},
	}
}

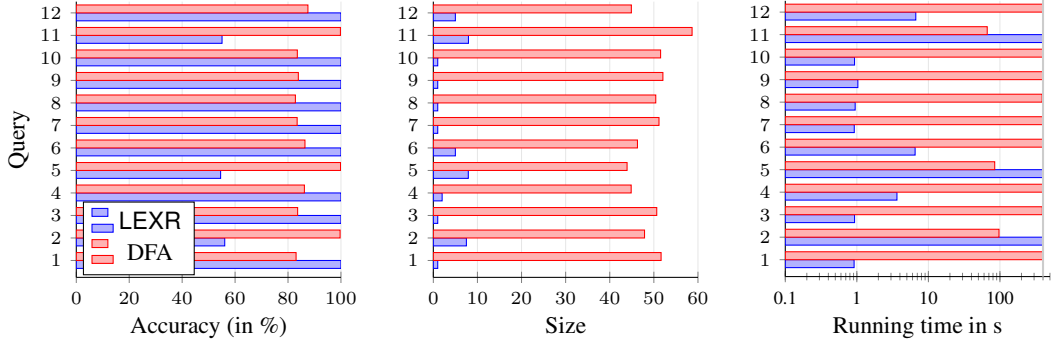
\begin{figure}[t]
	\centering
	
	\begin{comment}
	\begin{tikzpicture}
	\begin{axis}[barplot, width=74mm, height=45mm, ymin=0, xmin=0.25, xmax=12.5, xlabel={Query}, ylabel={Accuracy}, legend columns=2, legend style={at={(.975, .025)}, anchor=south east, font=\small, /tikz/every even column/.append style={column sep=0.5cm}}]
	\addplot[fill=gray!30] table [x=query-no, y=accuracy-LTL, col sep=comma] {csv/parenthesis.csv};
	\addlegendentry{\framework}
	\label{pgfplots:LEXR}
	\addplot[fill=gray!90] table [x=query-no, y=accuracy-DFA, col sep=comma] {csv/parenthesis.csv};
	\addlegendentry{DFA}
	\label{pgfplots:DFA}
	\end{axis}
	\end{tikzpicture}
	\hfill
	\begin{tikzpicture}
	\begin{axis}[barplot, width=74mm, height=45mm, ymin = 0, ymax=60, xmin=0.25, xmax=12.5, xlabel={Query}, ylabel={Size}]
	\addplot table [x=query-no, y=size-LTL, col sep=comma] {csv/parenthesis.csv};
	\addplot table [x=query-no, y=size-DFA, col sep=comma] {csv/parenthesis.csv};
	\end{axis}
	\end{tikzpicture}
	\end{comment}
	
	\begin{tikzpicture}
	\begin{axis}[barplot2, width=51mm, height=52.5mm, xmin=0, ymin=0.25, ymax=12.5, xlabel={Accuracy (in \%)}, ylabel={Query}, legend columns=1, legend style={at={(.025, .275)}, anchor=north west, font=\footnotesize}]
	\addplot table [x=accuracy-LTL, y=query-no, col sep=comma] {csv/parenthesis.csv};
	\addlegendentry{\framework}
	\addplot table [x=accuracy-DFA, y=query-no, col sep=comma] {csv/parenthesis.csv};
	\addlegendentry{DFA}
	\end{axis}
	\end{tikzpicture}
	\hfill
	\begin{tikzpicture}
	\begin{axis}[barplot2, width=51mm, height=52.5mm, xmin=0, xmax=60, ymin = 0.25, ymax=12.5, xtick distance=10, xlabel={Size}]
	\addplot table [x=size-LTL, y=query-no, col sep=comma] {csv/parenthesis.csv};
	\addplot table [x=size-DFA, y=query-no, col sep=comma] {csv/parenthesis.csv};
	\end{axis}
	\end{tikzpicture}
	\hfill
	\begin{tikzpicture}
	\begin{axis}[barplot2, width=51mm, height=52.5mm, xmin=.1, xmax=500, ymin = 0.25, ymax=12.5, xmode=log, log origin=infty, xlabel={Running time in s}, xtickten={-1,0, 1, 2}, xticklabels={0.1, 1, 10, 100}]
	\addplot table [x=time-LTL, y=query-no, col sep=comma] {csv/parenthesis.csv};
	\addplot table [x=time-DFA, y=query-no, col sep=comma] {csv/parenthesis.csv};
	\draw[color=gray!40, thick] (axis cs:400, .25) -- (axis cs:400, 12.75);
	\end{axis}
	\end{tikzpicture}
	\caption{Experimental results for twelve queries to a RNN for balanced parentheses.} 
	\label{fig:results}
	\label{fig:bp}
\end{figure}

The results of \framework\ and Mayr and Yovine's method are shown in Figure~\ref{fig:results}.
As can be seen from the bar plot on the left, \framework\ produces explanation with $100\%$ accuracy in nine out of twelve cases---in the remaining three, the average accuracy is still greater than $50\%$.
Moreover, for nine queries, the average accuracy of the LTL explanations was greater than the one of the DFAs.
Mayr and Yovine's method, on the other hand, was able to learn more accurate DFAs for three of the twelve queries.

The bar plot in the center of Figure~\ref{fig:results} shows the size of the generated explanations.
\framework's LTL explanations were relatively small.
The largest explanation was of size $8$.
On the other hand, the DFA explanations of Mayr and Yovine's method were comparatively large.
In fact, all DFAs had an average of $44$ or more states, making them arguably challenging to interpret.

Finally, the running times of both methods are shown in the right bar plot of Figure~\ref{fig:results}.
For three queries, \framework\ always times out and produces explanations with an average accuracy of $54\%$.
For the remaining nine queries, however, \framework\ is on average significantly faster than Mayr and Yovine's method (note the logarithmic axis) and produces explanations with $100\%$ accuracy.


\textbf{Summary~~}
Our experimental results show a similar pattern for all other RNNs (discussed in the Appendix).
In total, we observed that \framework\ is able to generate human-interpretable explanations for a variety of different RNN acceptors and queries.
Its explanations are comparable to the explanations of Mayr and Yovine's method in terms of average accuracy.
However, \framework's explanations are on average much smaller and, hence, easier to interpret.


\section{Conclusions}

In this paper, we have presented \framework, a novel explanation framework for RNNs based on the formal description language LTL.
In addition to the PAC-guarantee of \framework, we have empirically shown that \framework\ generates accurate explanations that are often smaller and, thus, easier to interpret than explanations produced by state-of-the-art methods that extract DFAs from RNN acceptors.
%
For future work, we plan to build a verifier based on conformance testing~\cite{fujiwara1991test,roth1966diagnosis}, which would allow learning an exact explanation for an RNN acceptor. 
Moreover, we plan to use a recent learner for PSL (Property Specification Language)~\cite{DBLP:journals/corr/abs-2002-03668} to increase the expressiveness of our explanations.

\section*{Appendix}


We have conducted experiments with six different RNN acceptors, three RNN acceptors for  synthetic problems and three RNN acceptors for real-word problems. For the synthetic problems, we have considered three languages $ \F(a) $, $ \F(a \wedge \X(b)) $, and $ \G(a \rightarrow \X(b)) $ over the alphabet $ \Sigma = \{a,b,c\} $, and for the real-word problems, we have studied balanced parentheses problem~\cite{DBLP:conf/icml/WeissGY18},  email pattern matching~\cite{DBLP:conf/icml/WeissGY18}, and alternating bit protocol~\cite{mayr2018regular}.
We have compared {\framework} with approximate DFA by Mayr and Yovine~\cite{mayr2018regular}.  For each experiment, we have repeated $ 250 $ times and shown the average accuracy, size of the  explanations and computation time. In addition, we have reported the explanation that has accuracy close to the average accuracy. In the following, we discuss all results in detail.

\subsection*{Synthetic Problem: $ \F(a) $ }

We have considered eleven queries when the RNN acceptor is trained on the language $ \F(a) $. In Table~\ref{tab:example0} we list all the queries and their  explanations, and in Figure~\ref{fig:results_F_a} we show the comparative average performance of LTL explanations with DFAs. 

\begin{table}[h]
	\caption{Example of LTL explanations when  the RNN is trained on the language 
		$ \F (a) $ 
		.}
	\label{tab:example0}

	\begin{center}
		\begin{tabular}{lrll}
			\toprule
			Problem & No. & Query & Explanation\\
			\midrule
			
			$ \F (a) $    
			& $ 1 $ & $ \F (a  \wedge  \X (b)) $ & $ \F  (a  \wedge  (\X  b)) $ \\ 
			\addlinespace[0em]
			& $ 2 $ & $ \F (a\U b) $ & $ (\F  a)  \wedge  (\F  b) $ \\ 
			\addlinespace[0em]
			& $ 3 $ & $ \F (b) $ & $ (\F  a)  \wedge  (\F  b) $ \\ 
			\addlinespace[0em]
			& $ 4 $ & $ \F (b\U a) $ & $ \F  a $ \\ 
			\addlinespace[0em]
			& $ 5 $ & $ \F (c) $ & $ (\F  a)  \wedge  (\F  c) $ \\ 
			\addlinespace[0em]
			& $ 6 $ & $ \F ( \neg a) $ & $ (\F  a)  \wedge  (a \U  ( \neg  a)) $ \\ 
			\addlinespace[0em]
			& $ 7 $ & $ \F ( \neg b) $ & $ \F  a $ \\ 
			\addlinespace[0em]
			& $ 8 $ & $ \G (a) $ & $ \G  a $ \\ 
			\addlinespace[0em]
			& $ 9 $ & $ \G (c) $ & $ \bot $ \\ 
			\addlinespace[0em]
			& $ 10 $ & $ \bot $ & $ \bot $ \\ 
			\addlinespace[0em]
			& $ 11 $ & $ \top $ & $ \F  a $ \\ 
			\addlinespace[0em]
			
			\bottomrule
		\end{tabular}
	\end{center} 
\end{table}

When the query is true ($ \top $), the learned explanation is $ \F(a) $ stating that the RNN can accurately learn the language it is supposed to learn. We have considered another query $ \F(a\U b) $ that accepts a word where eventually $ a $ appears until $ b $ appears. For this query, {\framework} learns an explanation $ \F(a) \wedge (\F b) $, which is in fact the minimal equivalent of  
$ \F(a) \wedge \F(a \U b) $. The other queries and their explanations can also be interpreted in the similar manner.  We next discuss the performance of LTL explanations compared against DFAs.

\pgfplotsset{
	barplot2/.style={
		xbar=0pt,
		bar width=3pt,
		ytick distance=1,
		tick label style={font=\scriptsize},
		tick align=outside,
		xlabel style={font=\footnotesize\strut, yshift=1mm},
		ylabel style={font=\footnotesize, yshift=-1mm},
		axis x line=bottom,
		axis y line=left,
		axis line style={-},
		xmajorgrids=true,
		major grid style={color=gray!20, very thin},
	}
}

\begin{figure}[h]
	\centering

	\begin{tikzpicture}
	\begin{axis}[barplot2, width=51mm, height=50.5mm, xmin=0.01, ymin=0.25, ymax=11.5, xlabel={Accuracy (in \%)}, ylabel={Query}, legend columns=1, legend style={at={(.025, .295)}, anchor=north west, font=\footnotesize}]
	\addplot table [x=accuracy-LTL, y=query-no, col sep=comma] {csv/F-a.csv};
	\addlegendentry{\framework}
	\addplot table [x=accuracy-DFA, y=query-no, col sep=comma] {csv/F-a.csv};
	\addlegendentry{DFA}
	\end{axis}
	\end{tikzpicture}
	\hfill
	\begin{tikzpicture}
	\begin{axis}[barplot2, width=51mm, height=50.5mm, xmin=.8, xmax=100, ymin = 0.25, ymax=11.5, xlabel={Size},xmode=log, log origin=infty, xtickten={-1,0, 1, 2}, xticklabels={0, 1, 10, 100}]
	\addplot table [x=size-LTL, y=query-no, col sep=comma] {csv/F-a.csv};
	\addplot table [x=size-DFA, y=query-no, col sep=comma] {csv/F-a.csv};
	\end{axis}
	\end{tikzpicture}
	\hfill
	\begin{tikzpicture}
	\begin{axis}[barplot2, width=51mm, height=50.5mm, xmin=.1, xmax=500, ymin = 0.25, ymax=11.5, xmode=log, log origin=infty, xlabel={Running time in s}, xtickten={-1,0, 1, 2}, xticklabels={0.1, 1, 10, 100}]
	\addplot table [x=time-LTL, y=query-no, col sep=comma] {csv/F-a.csv};
	\addplot table [x=time-DFA, y=query-no, col sep=comma] {csv/F-a.csv};
	\draw[color=gray!40, thick] (axis cs:400, .25) -- (axis cs:400, 12.75);
	\end{axis}
	\end{tikzpicture}
		\caption{Experimental results for eleven queries to a RNN for the language  $ \F(a) $. Explanation size and running time are in the log scale.} 
	\label{fig:results_F_a}
\end{figure}
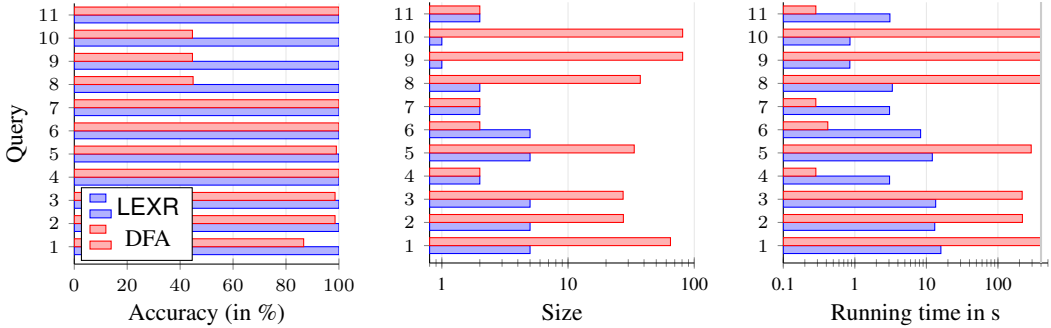

As can be seen from the leftmost bar plot in Figure~\ref{fig:results_F_a}, the learned LTL explanations have $ 100\% $ average accuracy in all eleven queries. In contrast, DFA has more than $  98 \% $ average  accuracy in seven queries. In addition, DFA has $ 45 \% $ average accuracy in three queries  where the LTL explanation is false for two queries  and $ \G(a) $ for the third query.  From this observation we have hypothesized that there exists languages that are learned efficiently as an LTL but not as a DFA.

In the middle bar plot in Figure~\ref{fig:results_F_a}, we show the average  size of the two explanations. In general, we have observed that the DFA has more average size in majority of the queries, and it is particularly worse where the average accuracy of the DFAs is less than that of the LTL explanations. For example, when the query is $ \G(c) $, the LTL explanation with average size $ 1 $ has $ 100\% $ accuracy whereas the DFA has on average $ 80 $ states with $ 45 \% $ accuracy.  Therefore, the LTL explanations are not only more accurate but also smaller in size. 

In the rightmost bar plot, we show the average computation time of the compared explanations where the DFA learner costs higher computation time than the LTL learner in seven out of eleven queries. Among these seven queries, the DFA learner times out in four queries. In this context, the average computation time of {\framework} is at most $ 16 $ seconds in all reported queries. 
Therefore, from the above analysis for the synthetic problem $ \F(a) $ it is clear that the LTL explanations are generally better than the DFAs in all three aspects:  accuracy, size and computation time.

\subsection*{Synthetic Problem: $ \F(a \wedge \X(b)) $}

We have trained a RNN acceptor on the language $ \F(a \wedge \X(b)) $ that accepts a word containing the infix $ ab $. For this problem, we  have considered seven queries listed in Table~\ref{tab:example1}. When the query considers words containing only symbol $ a $, i.e., the query is $ \G(a) $, it is evident that there is no accepting word satisfying  the query and the language of the RNN. Therefore, the LTL explanation is false. We have discussed the query $ \F(c) $ in the main paper where the LTL learner cannot learn a PAC explanation before $ 400 $ seconds. While the learned explanation $ (\F  (\X  b))  \wedge  (a \U  ( \neg  (\X  b))) $ has size $ 7 $, the expected minimal explanation is $ \F(c) \wedge \F(a \wedge \X(b)) $ with size $ 8 $. As a result, we have hypothesized that if more computational time is allotted, {\framework} would learn the expected explanation provided that the RNN is perfect. The other queries in Table ~\ref{tab:example1} are self-explanatory and thereby we discuss the comparative performance analysis of {\framework} with the work of Mayr and Yovine in the following.

\begin{table}[h]
	\caption{Example of LTL explanations of the RNN on the language		$ \F (a  \wedge  \X (b)) $.}
	\label{tab:example1}

	\begin{center}
		\begin{tabular}{lrll}
			\toprule
			Problem & No. & Query & Explanation\\
			\midrule
			
			$ \F (a  \wedge  \X (b)) $    
			& $ 1 $ & $ \F (a) $ & $ \F  (a  \wedge  (\X  b)) $ \\ 
			\addlinespace[0em]
			& $ 2 $ & $ \F (a\U b) $ & $ \F  (a  \wedge  (\X  b)) $ \\ 
			\addlinespace[0em]
			& $ 3 $ & $ \F (b) $ & $ \F  (a  \wedge  (\X  b)) $ \\ 
			\addlinespace[0em]
			& $ 4 $ & $ \F (c) $ & $ (\F  (\X  b))  \wedge  (a \U  ( \neg  (\X  b))) $ \\ 
			\addlinespace[0em]
			& $ 5 $ & $ \G (a) $ & $ \bot $ \\ 
			\addlinespace[0em]
			& $ 6 $ & $ \bot $ & $ \bot $ \\ 
			\addlinespace[0em]
			& $ 7 $ & $ \top $ & $ \F  (a  \wedge  (\X  b)) $ \\ 
			\addlinespace[0em]
			
			\bottomrule
		\end{tabular}
	\end{center} 
\end{table} 

\begin{figure}[h]
	\centering

	\begin{tikzpicture}
	\begin{axis}[barplot2, width=51mm, height=40.5mm, xmin=0, ymin=0.25, ymax=7.5, xlabel={Accuracy (in \%)}, ylabel={Query}, legend columns=1, legend style={at={(.025, .475)}, anchor=north west, font=\footnotesize}]
	\addplot table [x=accuracy-LTL, y=query-no, col sep=comma] {csv/F-a-x-b.csv};
	\addlegendentry{\framework}
	\addplot table [x=accuracy-DFA, y=query-no, col sep=comma] {csv/F-a-x-b.csv};
	\addlegendentry{DFA}
	\end{axis}
	\end{tikzpicture}
	\hfill
	\begin{tikzpicture}
	\begin{axis}[barplot2, width=51mm, height=40.5mm, xmin=0, xmax=50, ymin = 0.25, ymax=7.5, xtick distance=10, xlabel={Size}]
	\addplot table [x=size-LTL, y=query-no, col sep=comma] {csv/F-a-x-b.csv};
	\addplot table [x=size-DFA, y=query-no, col sep=comma] {csv/F-a-x-b.csv};
	\end{axis}
	\end{tikzpicture}
	\hfill
	\begin{tikzpicture}
	\begin{axis}[barplot2, width=51mm, height=40.5mm, xmin=.1, xmax=500, ymin = 0.25, ymax=7.5, xmode=log, log origin=infty, xlabel={Running time in s}, xtickten={-1,0, 1, 2}, xticklabels={0.1, 1, 10, 100}]
	\addplot table [x=time-LTL, y=query-no, col sep=comma] {csv/F-a-x-b.csv};
	\addplot table [x=time-DFA, y=query-no, col sep=comma] {csv/F-a-x-b.csv};
	\draw[color=gray!40, thick] (axis cs:400, .25) -- (axis cs:400, 12.75);
	\end{axis}
	\end{tikzpicture}
	\caption{Experimental results for seven queries to a RNN for the language $ \F(a \wedge \X(b)) $.  Running time is in the log scale.} 
	\label{fig:results_F_a_x_b}
\end{figure}

In Figure~\ref{fig:results_F_a_x_b}, we present the average accuracy of explanations in the left plot, the average size of explanations in the middle plot, and the average computation time in the right plot. For all queries except the fourth query, both LTL and DFA have similar average performance: all explanations have $ 100\% $  accuracy and small size (less than $ 5 $) and cost less computation time (less than $ 16 $ seconds). For the fourth query $ \F(c) $, both LTL and DFA learner time out where the extracted DFA is on average larger ($ 44 $ states)  than the LTL ($ 7 $ sub-formulas) but more accurate ($ 100\% $) than the LTL ($ 57\% $).


\subsection*{Synthetic Problem $ \G(a \rightarrow \X(b)) $}

For the last synthetic problem, we have trained a RNN acceptor on the language $ \G(a \rightarrow \X(b)) $ that accepts a word where every $ a $ is immediately followed by $ b $. Compared to the previous two synthetic problems, this language accepts a smaller region of the input space. 

For this problem, we have considered seven queries in Table~\ref{tab:example2}. We here discuss Query $ 1 $ and $ 7 $ as the other queries and their explanations can be interpreted trivially. When the query is true, {\framework} learns a minimal size and PAC explanation $ \G  ((\X  (c \U  b)) \U  (b \vee  c)) $ with size $ 7 $. However, the expected LTL explanation is  $ \G(a \rightarrow \X(b)) $ with size $ 5 $ and hence, {\framework} certifies that the RNN has not learned the exact language it is supposed to learn in spite of showing $ 100\% $ accuracy on the test set. In fact, the verifier in {\framework} finds that there is a word $ abcabacbccbbbabcbcbccbbccbbcccccc $ that is not satisfied by $ \G(a \rightarrow \X(b)) $ but accepted by the RNN. For the same reason of the RNN not capturing the language $ \G(a \rightarrow \X(b)) $ perfectly, when we have considered a query $ \F(a) $,  the learned explanation is different from the expected minimal explanation $ \G(a \rightarrow \X(b)) \wedge \F(a) $. 

\begin{table}[h]
	\caption{Example of LTL explanations of the RNN on
	the language 	$ \G (a \rightarrow\X (b)) $.}
	\label{tab:example2}

	\begin{center}
		\begin{tabular}{lrll}
			\toprule
			Problem & No. & Query & Explanation\\
			\midrule
			
			$ \G (a \rightarrow\X (b)) $    
			& $ 1 $ & $ \F (a) $ & $ (\G  (a  \rightarrow (a \U  b))) \U  (a  \wedge  (\G  (a  \rightarrow (a \U  b)))) $ \\ 
			\addlinespace[0em]
			& $ 2 $ & $ \G (a) $ & $ \bot $ \\ 
			\addlinespace[0em]
			& $ 3 $ & $ \G (b) $ & $ \G  b $ \\ 
			\addlinespace[0em]
			& $ 4 $ & $ \X (b) $ & $ (\X  b)  \wedge  (\G  (a  \rightarrow (\X  b))) $ \\ 
			\addlinespace[0em]
			& $ 5 $ & $ b $ & $ b  \wedge  (\G  (a  \rightarrow (\X  b))) $ \\ 
			\addlinespace[0em]
			& $ 6 $ & $ \bot $ & $ \bot $ \\ 
			\addlinespace[0em]
			& $ 7 $ & $ \top $ & $ \G  ((\X  (c \U  b)) \U  (b \vee  c)) $ \\ 
			\addlinespace[0em]
			
			\bottomrule
		\end{tabular}
	\end{center} 
\end{table} 

\begin{figure}[h]
	\centering

	\begin{tikzpicture}
	\begin{axis}[barplot2, width=51mm, height=40.5mm, xmin=0, ymin=0.25, ymax=7.5, xlabel={Accuracy (in \%)}, ylabel={Query}, legend columns=1, legend style={at={(.025, .475)}, anchor=north west, font=\footnotesize}]
	\addplot table [x=accuracy-LTL, y=query-no, col sep=comma] {csv/G-a-X-b.csv};
	\addlegendentry{\framework}
	\addplot table [x=accuracy-DFA, y=query-no, col sep=comma] {csv/G-a-X-b.csv};
	\addlegendentry{DFA}
	\end{axis}
	\end{tikzpicture}
	\hfill
	\begin{tikzpicture}
	\begin{axis}[barplot2, width=51mm, height=40.5mm, xmin=.8, xmax=1000, ymin = 0.25, ymax=7.5, 
	 xlabel={Size},xmode=log, log origin=infty,xtickten={-1,0, 1, 2, 2.69}, xticklabels={0, 1, 10, 100,500}]
	\addplot table [x=size-LTL, y=query-no, col sep=comma] {csv/G-a-X-b.csv};
	\addplot table [x=size-DFA, y=query-no, col sep=comma] {csv/G-a-X-b.csv};
	\end{axis}
	\end{tikzpicture}
	\hfill
	\begin{tikzpicture}
	\begin{axis}[barplot2, width=51mm, height=40.5mm, xmin=.1, xmax=500, ymin = 0.25, ymax=7.5, xmode=log, log origin=infty, xlabel={Running time in s}, xtickten={-1,0, 1, 2}, xticklabels={0.1, 1, 10, 100}]
	\addplot table [x=time-LTL, y=query-no, col sep=comma] {csv/G-a-X-b.csv};
	\addplot table [x=time-DFA, y=query-no, col sep=comma] {csv/G-a-X-b.csv};
	\draw[color=gray!40, thick] (axis cs:400, .25) -- (axis cs:400, 12.75);
	\end{axis}
	\end{tikzpicture}
	\caption{Experimental results for seven queries to a RNN for the language  $ \G(a\rightarrow \X(b)) $. Explanation size and running time are in the log scale.} 
	
	\label{fig:results_G_a_X_b}
\end{figure}
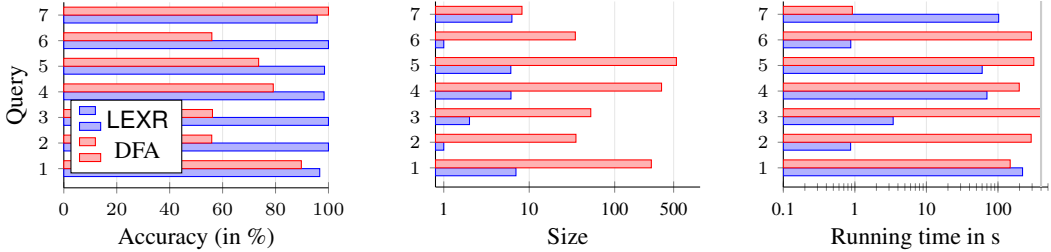

In Figure~\ref{fig:results_G_a_X_b}, we have presented the comparative average performance of LTL explanations and DFAs. In the left plot, LTL explanations have higher accuracy than DFAs in all queries expect the query $ \top $. In fact, all LTL explanations are PAC explanations of the RNN certified by the same verifier that both LTL and DFA learners call. But this is not the case for the DFA learner as it times out in query $ 3 $  as shown in the rightmost plot. In addition, we have observed that the DFA learner incurs higher computational cost on average than the LTL learner in majority of the queries.  Moving focus to the size of the explanations in the middle plot,  LTL explanations are much smaller than DFAs in most queries. Therefore, LTL explanations have outperformed the DFAs in all three aspects as the RNN acceptor is trained on the language $ \G(a \rightarrow \X(b)) $.

\subsection*{Balanced Parentheses Problem (BP)}
\begin{figure}[h]
	\centering

	\begin{tikzpicture}
	\begin{axis}[barplot2, width=51mm, height=52.5mm, xmin=0, ymin=0.25, ymax=12.5, xlabel={Accuracy (in \%)}, ylabel={Query}, legend columns=1, legend style={at={(.025, .275)}, anchor=north west, font=\footnotesize}]
	\addplot table [x=accuracy-LTL, y=query-no, col sep=comma] {csv/parenthesis.csv};
	\addlegendentry{\framework}
	\addplot table [x=accuracy-DFA, y=query-no, col sep=comma] {csv/parenthesis.csv};
	\addlegendentry{DFA}
	\end{axis}
	\end{tikzpicture}
	\hfill
	\begin{tikzpicture}
	\begin{axis}[barplot2, width=51mm, height=52.5mm, xmin=0, xmax=60, ymin = 0.25, ymax=12.5, xtick distance=10, xlabel={Size}]
	\addplot table [x=size-LTL, y=query-no, col sep=comma] {csv/parenthesis.csv};
	\addplot table [x=size-DFA, y=query-no, col sep=comma] {csv/parenthesis.csv};
	\end{axis}
	\end{tikzpicture}
	\hfill
	\begin{tikzpicture}
	\begin{axis}[barplot2, width=51mm, height=52.5mm, xmin=.1, xmax=500, ymin = 0.25, ymax=12.5, xmode=log, log origin=infty, xlabel={Running time in s}, xtickten={-1,0, 1, 2}, xticklabels={0.1, 1, 10, 100}]
	\addplot table [x=time-LTL, y=query-no, col sep=comma] {csv/parenthesis.csv};
	\addplot table [x=time-DFA, y=query-no, col sep=comma] {csv/parenthesis.csv};
	\draw[color=gray!40, thick] (axis cs:400, .25) -- (axis cs:400, 12.75);
	\end{axis}
	\end{tikzpicture}
		\caption{Experimental results for twelve queries to a RNN for balanced parentheses problem. Running time is in the log scale.} 
	\label{fig:results_bp}
\end{figure}
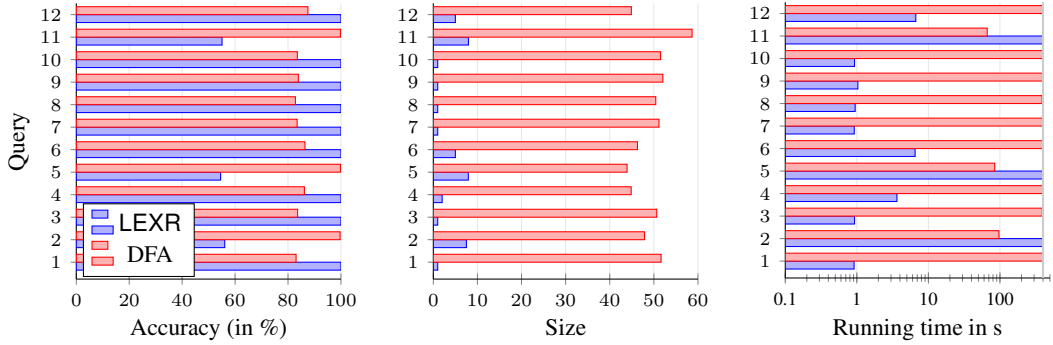

For this benchmark, the RNN acceptor is trained to predict whether a word has balanced parentheses or not.
We have used the alphabet $\Sigma = \{ l, r, a \}$ where $l$ and $r$ are placeholders for the left parenthesis ``('' and the right parenthesis ``)'', respectively, and $a$ stands for non-parenthesis symbols.
For our evaluation, we have considered twelve queries in Table~\ref{tab:example3}, which define different valid and invalid properties of balanced parentheses.  In the main paper, we have discussed the comparative performance analysis between two explanations as shown in Figure~\ref{fig:results_bp} and hence do not repeat here. Instead, we discuss the learned explanations in the following.

\begin{table}
	\caption{Example of LTL explanations of the RNN on balanced parentheses problem.}
	\label{tab:example3}

	\begin{center}
		\begin{tabular}{lrll}
			\toprule
			Problem & No. & Query & Explanation\\
			\midrule
			
			BP   
			& $ 1 $ & $ \F (l  \wedge  \X (\G ( \neg r))) $ & $ \bot $ \\ 
			\addlinespace[0em]
			& $ 2 $ & $ \F (l)  \wedge  \F (r)  \wedge  \F ( (l\vee a) \U  r ) $ & $ ((a \U  r)  \rightarrow r)  \wedge  (a \U  (\X  (a \U  r))) $ \\ 
			\addlinespace[0em]
			& $ 3 $ & $ \F (l)  \wedge  \F (r)  \wedge   \neg (\F ( (l\vee a) \U  r )) $ & $ \bot $ \\ 
			\addlinespace[0em]
			& $ 4 $ & $ \G (a) $ & $ \G  a $ \\ 
			\addlinespace[0em]
			& $ 5 $ & $ \G (l  \rightarrow \F (r)) $ & $ ((\F  l) \U  ( \neg  (\F  l)))  \wedge  ( \neg  (( \neg  (\F  l)) \U  r)) $ \\ 
			\addlinespace[0em]
			& $ 6 $ & $ \G (l  \rightarrow  \neg (\F (r))) $ & $ \G  ( \neg  (l \vee  r)) $ \\ 
			\addlinespace[0em]
			& $ 7 $ & $ \G (l) $ & $ \bot $ \\ 
			\addlinespace[0em]
			& $ 8 $ & $ a \U  r $ & $ \bot $ \\ 
			\addlinespace[0em]
			& $ 9 $ & $ \bot $ & $ \bot $ \\ 
			\addlinespace[0em]
			& $ 10 $ & $ r $ & $ \bot $ \\ 
			\addlinespace[0em]
			& $ 11 $ & $ \top $ & $ ((\F  l) \U  ( \neg  (\F  l)))  \wedge  ( \neg  (( \neg  (\F  l)) \U  r)) $ \\ 
			\addlinespace[0em]
			& $ 12 $ & $  \neg \F (l\vee r) $ & $ \G  ( \neg  (l \vee  r)) $ \\ 
			\addlinespace[0em]
			
			\bottomrule
		\end{tabular}
	\end{center} 
\end{table} 

The query $ \F (l  \wedge  \X (\G ( \neg r))) $  accepts a word where eventually there is a left parenthesis and from the next position onwards there is no right parenthesis. {\framework} learns a false ($ \bot $) explanation stating that there is no valid balanced parentheses satisfying the query. 
Similarly, we have considered simple queries $ \G(l), a \U r, r $ as  sanity checks for the RNN such that the queries accept a word where the left parenthesis is in every position, non parenthesis symbol appears until there is a right parenthesis, and the word starts with a right parenthesis, respectively. {\framework} learns an explanation $ \bot $ for all the three queries certifying that the RNN passes the sanity checks. 

If the word contains non-parenthesis symbols only, the RNN should accept the word. Hence we have considered a query $ \neg \F(l \vee r) $ and the learned explanation is $ \G(\neg (l \vee r)) $, i.e, indeed the RNN accepts a word where in every position there is a non-parenthesis symbol. 

In all above mentioned queries, the LTL explanations are PAC explanations of the RNN. We now discuss a query, where the explanation is not a PAC explanation. We have considered a query $ \G( l \rightarrow \F(r)) $: in every position of the word if there is a left parenthesis then eventually there is a right parenthesis. For this query, {\framework} times out and  learns an LTL formula $  \varphi =   ((\F  l) \U  ( \neg  (\F  l)))  \wedge  ( \neg  (( \neg  (\F  l)) \U  r))  $. This explanation states that a word is valid parentheses if  in the word, (i) eventually left parenthesis appears until in every position onwards left parenthesis does not appear and (ii) it is not the case that the left parenthesis never appears  until there is a right parenthesis. As the explanation for this query does not have the PAC guarantee,  we cannot exactly infer what the RNN has  learned. Hence we have considered a query $ \G(l \rightarrow \neg \F(r)) $: in every position of the accepting word, if there is a left parenthesis then there is never a right parenthesis. {\framework} learns an explanation $ \G(\neg (l \vee r)) $ for this query stating that the RNN inside the query only accepts a word if it contains no parenthesis. Intuitively,  as in the query every left parenthesis is immediately followed by any symbol other than the right parenthesis, an accepting word of the RNN cannot have any parenthesis at all.

\subsection*{Email Pattern Matching}
In the email pattern matching problem,  a valid email address is defined by the following regular expression. 
\begin{equation*}
\label{eq:reg-email}
\text{[a-z][a-z0-9]\textasteriskcentered @[a-z0-9]+.[a-z]+} \$
\end{equation*}
In this regular expression,  {\textasteriskcentered } refers to the Kleene star operation indicating zero or more occurrences of the preceding element, $ + $ indicates one or more occurrences of the preceding element, and $ \$ $  matches the end position of the word.
In our prototype implementation, we have considered an alphabet $ \{p,m,@,\circ\} $  with the following abstraction: $ p $ corresponds to the class of symbols `[a-z]', $ m $ corresponds to the class of symbols `[0-9]', $ \circ $ is the placeholder for  `.', and $ @ $ is the placeholder for  `@'.  
We have considered a set of nineteen queries by specifying different properties of valid and invalid email addresses and asked for an explanation to see if the RNN has learned them accurately. The queries and their explanations are presented in Table~\ref{tab:example4}, which we discuss next.

\begin{table}[h]
	\caption{Example of LTL explanations of the RNN on
		Email pattern matching problem.}
	\label{tab:example4}

	\begin{center}
		\begin{tabular}{lrp{5cm}l}
			\toprule
			Problem & No. & Query & Explanation\\
			\midrule
			
			Email   
			& $ 1 $ & $ m $ & $ \X  (\X  (\X  (\X  (\X  \circ )))) $ \\ 
			\addlinespace[0em]
			& $ 2 $ & $ ( \neg \F (m))  \wedge  \F (p \U  @)  \wedge  \F (@  \wedge  \X (p \U  \circ ))  \wedge  \F (\circ   \wedge  \X (\G p)) $ & $ p \U  (\X  (@  \wedge  (\X  p))) $ \\ 
			\addlinespace[0em]
			& $ 3 $ & $ \F ((p\vee m) \U  @)  \wedge  \F (@  \wedge  \X ((p\vee m) \U  \circ ))  \wedge  \F (\circ   \wedge  \X (\G p)) $ & $ \X  (\X  (\X  (\X  ( \neg  @)))) $ \\ 
			\addlinespace[0em]
			& $ 4 $ & $ \F (@  \wedge  \X (\F @)) $ & $ \X  (\X  (\X  (\X  (\X  (\X  p))))) $ \\ 
			\addlinespace[0em]
			& $ 5 $ & $ \F (@  \wedge  \X (\G ( \neg \circ ))) $ & $ \X  (\X  (\X  (\X  (\X  p)))) $ \\ 
			\addlinespace[0em]
			& $ 6 $ & $ \F (@  \wedge  \X (\circ )) $ & $ \bot $ \\ 
			\addlinespace[0em]
			& $ 7 $ & $ \F (\circ   \wedge  \X (\F \circ )) $ & $ \X  ((\X  (\X  p)) \U  (p  \wedge  (\X  (\X  p)))) $ \\ 
			\addlinespace[0em]
			& $ 8 $ & $ \F (\circ   \wedge  \X (\F m)) $ & $ (\F  m)  \wedge  (\X  (\X  p)) $ \\ 
			\addlinespace[0em]
			& $ 9 $ & $ \F (\circ   \wedge  \X (\G p)) $ & $ \X  ((\F  p) \U  (\X  (\circ   \wedge  (\F  p)))) $ \\ 
			\addlinespace[0em]
			& $ 10 $ & $ \G (m) $ & $ \bot $ \\ 
			\addlinespace[0em]
			& $ 11 $ & $ @ $ & $ \bot $ \\ 
			\addlinespace[0em]
			& $ 12 $ & $ \circ  $ & $ \bot $ \\ 
			\addlinespace[0em]
			& $ 13 $ & $ \bot $ & $ \bot $ \\ 
			\addlinespace[0em]
			& $ 14 $ & $ \top $ & $ \F  (p  \wedge  (\F  (\X  (\X  (\X  p))))) $ \\ 
			\addlinespace[0em]
			& $ 15 $ & $  \neg \F (@) $ & $ \X  ((m  \wedge  (\X  m)) \vee  (\X  (\X  m))) $ \\ 
			\addlinespace[0em]
			& $ 16 $ & $  \neg \F (\circ ) $ & $ (m  \rightarrow (\X  (\X  m)))  \wedge  (\X  (\X  (\X  m))) $ \\ 
			\addlinespace[0em]
			& $ 17 $ & $  \neg \F (m) $ & $ \X  ((\X  p)  \wedge  (\X  (\X  (\X  p)))) $ \\ 
			\addlinespace[0em]
			& $ 18 $ & $  \neg \F (p) $ & $ \bot $ \\ 
			\addlinespace[0em]
			& $ 19 $ & $  \neg p $ & $ m  \wedge  (\X  (\X  (\X  \top))) $ \\ 
			\addlinespace[0em]
			
			\bottomrule
		\end{tabular}
	\end{center} 
\end{table}

According to the regular expression of an email address, a valid  address cannot start with  digits `[0-9]'. Hence we have considered a query $ m $ that accepts words beginning with digits. The expected explanation for this query is false, however, {\framework} learns an explanation $ \X  (\X  (\X  (\X  (\X  \circ )))) $ with an interpretation that if a word starts with digits, then the RNN accepts it  if  there is the symbol `.' in the fifth position of the word. The reason for learning such an explanation is that the verifier in {\framework} has found multiple counterexamples starting with digits but still being accepted by the RNN. 
Therefore, {\framework} certifies that the RNN has not learned the property specified by the query correctly, although it has shown high accuracy ($ 99.97\% $ ) on a test set of around $ 3,800 $ random words. We have found similar misbehavior of the RNN on several other queries. For instance, the query $ \F (@  \wedge  \X (\F @)) $ accepts a word where at least two `@'s are present in a valid email address and the query $  \F (@  \wedge  \X (\G ( \neg \circ )))  $ accepts a word where eventually `@' appears and from the next position onwards there is no `.'. For both  queries, the RNN has accepted few words which it should not have accepted if trained accurately. Therefore, {\framework} shows the promise that our technology can be used for ``verification'' purposes. 
 
The RNN acceptor in the email pattern match problem succeeds in several sanity checks which we discuss now.  A valid email address cannot start with `@' or `.' and we have considered two queries $ @ $ and $ \circ $, respectively. For both queries, the learned (and expected) explanation is false. In addition, if the word contains only digits, i.e, the query is $ \G(m) $, the learned (and expected) explanation is false.

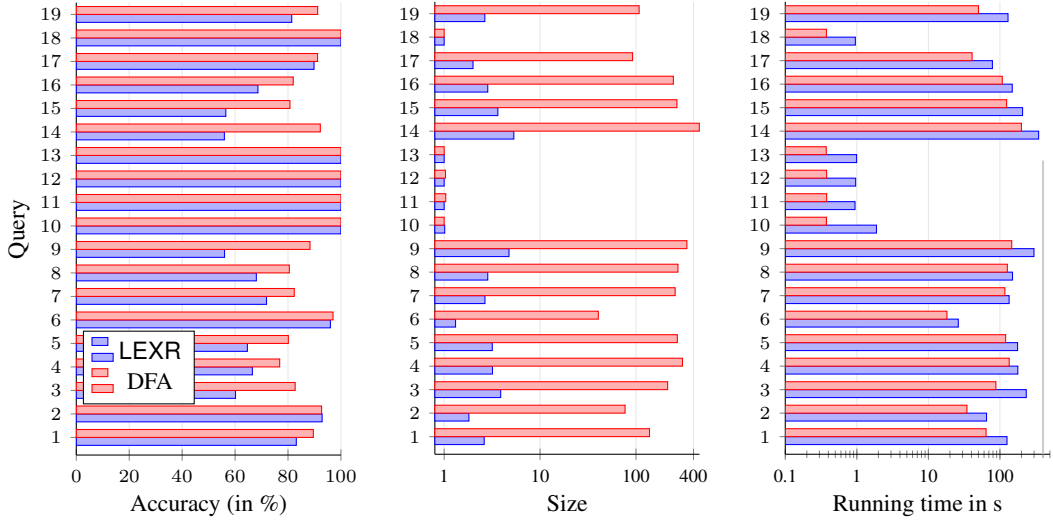
\begin{figure}[h]
	\centering

	\begin{tikzpicture}
	\begin{axis}[barplot2, width=51mm, height=76mm, xmin=0, ymin=0.25, ymax=19.5, xlabel={Accuracy (in \%)}, ylabel={Query}, legend columns=1, legend style={at={(.025, .275)}, anchor=north west, font=\footnotesize}]
	\addplot table [x=accuracy-LTL, y=query-no, col sep=comma] {csv/email.csv};
	\addlegendentry{\framework}
	\addplot table [x=accuracy-DFA, y=query-no, col sep=comma] {csv/email.csv};
	\addlegendentry{DFA}
	\end{axis}
	\end{tikzpicture}
	\hfill
	\begin{tikzpicture}
	\begin{axis}[barplot2, width=51mm, height=76mm, xmin=0.8, xmax=460, ymin = 0.25, ymax=19.5, xtick distance=100, xlabel={Size}, xmode=log, log origin=infty,xtickten={-1,0, 1, 2, 2.6}, xticklabels={0, 1, 10, 100,400}]
	\addplot table [x=size-LTL, y=query-no, col sep=comma] {csv/email.csv};
	\addplot table [x=size-DFA, y=query-no, col sep=comma] {csv/email.csv};
	\end{axis}
	\end{tikzpicture}
	\hfill
	\begin{tikzpicture}
	\begin{axis}[barplot2, width=51mm, height=76mm, xmin=.1, xmax=500, ymin = 0.25, ymax=19.5, xmode=log, log origin=infty, xlabel={Running time in s}, xtickten={-1,0, 1, 2}, xticklabels={0.1, 1, 10, 100}]
	\addplot table [x=time-LTL, y=query-no, col sep=comma] {csv/email.csv};
	\addplot table [x=time-DFA, y=query-no, col sep=comma] {csv/email.csv};
	\draw[color=gray!40, thick] (axis cs:400, .25) -- (axis cs:400, 12.75);
	\end{axis}
	\end{tikzpicture}
	\caption{Experimental results for nineteen queries to a RNN for email pattern match problem. Explanation size and running time are in the log scale.} 
	
	\label{fig:results_email}
\end{figure}

We now discuss queries where the expected explanation is not false. For instance, we have made an assumption of the language of the RNN in a restricted setting and  considered a query $ ( \neg \F (m))  \wedge  \F (p \U  @)  \wedge  \F (@  \wedge  \X (p \U  \circ ))  \wedge  \F (\circ   \wedge  \X (\G p)) $. This query is inspired by the regular expression of the valid email address  and accepts a word if in the word (i) there is no digit, (ii) eventually letters appear until `@' appears, (iii) eventually `@' appears and from the next position onwards letters appear until `.' appears, and (iv) eventually `.' appears and from the next position onwards letter always appear. {\framework} then learns an LTL formula $ p \U  (\X  (@  \wedge  (\X  p)))  $ that accepts words where `[a-z]' appear until the infix `@[a-z]' at the next position. In this context, note that this explanation accepts words that additionally satisfy the query. Therefore, the LTL explanation imposes more constraints  by specifying the relative position of letters and `@'.  We next discuss the comparative performance of LTL explanations with DFAs as presented in Figure~\ref{fig:results_email}.

We have presented the average accuracy of the two explanations LTL and DFA in the left plot in Figure~\ref{fig:results_email}. As we have discussed earlier, the RNN acceptor has not behaved as expected in several queries (query no. $ 1,4,5,7,8,15,16,  19 $ etc.) and the LTL explanations have comparatively less accuracy than that of  DFAs in those queries. However,  the DFAs are very large (more than $ 100 $ states on average)  as shown in the middle plot. Therefore, even if DFAs generalize well, they are not easy to understand. On the contrary, a small LTL explanation can indicate such anomaly in an interpretable manner. In  other queries, LTL formulas are competitive to DFAs in terms of average accuracy and better than DFAs in terms of the size of the explanations. We present the average computation time of the two explanations in the rightmost plot where the LTL learner generally takes  slightly more time to generate explanations than the DFA learner.

\subsection*{Alternating Bit Protocol.}
\label{sec:abp}
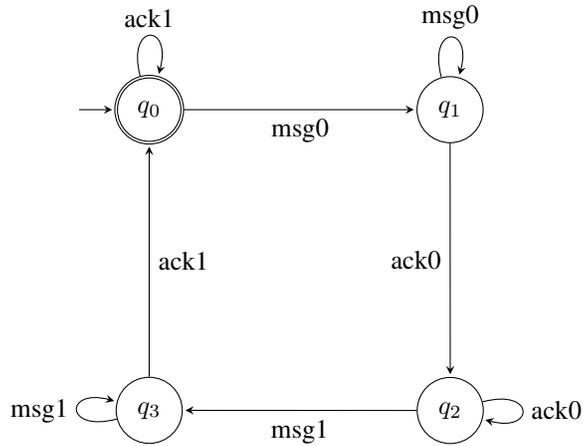
\begin{figure}
	\centering
	\begin{tikzpicture}[shorten >=1pt,node distance=4cm,on grid,auto, initial text = \textcolor{red}{}] 
	\node[state,initial,accepting] (q_0)   {$q_0$}; 
	\node[state] (q_1) [right=of q_0] {$q_1$}; 
	\node[state] (q_3) [below=of q_0] {$q_3$}; 
	\node[state](q_2) [below right=5.66 cm of q_0] {$q_2$};
	
	\path[->] 
	(q_0) edge  [loop above] node {ack1} (q_0)
	edge  node [swap] {msg0} (q_1)
	(q_1) edge  [loop above]  node  {msg0} (q_1)
	edge [left] node {ack0} (q_2)
	(q_2) edge  node  {msg1} (q_3) 
	edge [loop right] node {ack0} ()
	(q_3) edge  node [swap] {ack1} (q_0) 
	edge [loop left] node  {msg1} ()
	;
	\end{tikzpicture}
	\caption{Deterministic finite automata (DFA) for alternating bit protocol.}
	\label{fig:alternating-bit-protocol}
	
\end{figure}

In the last real-world benchmarks, we have studied alternating bit protocol  that is represented as a DFA in Figure~\ref{fig:alternating-bit-protocol} over the   alphabet $ \{\text{msg}0,\text{ack}0,\text{msg}1,\text{ack}1\} $.  We have trained a RNN acceptor to recognize this automata and
designed ten queries listed in Table~\ref{tab:example5} to evaluate the explanations on this benchmark. In the following, we discuss the queries and their explanations. 

\begin{table}[h]
	\caption{Example of LTL explanations of the RNN on
		alternating bit protocol.}
	\label{tab:example5}

	\begin{center}
		\begin{tabular}{lrp{5 cm}p{5 cm}}
			\toprule
			Problem & No. & Query & Explanation\\
			\midrule
			
			Bit   
			& $ 1 $ & $ \F (\text{msg}0 \U  \text{ack}0)  \wedge  \F (\text{ack}0 \U  \text{msg}1)  \wedge  \F (\text{msg}1 \U  \text{ack}1)  $ & $ \text{msg}0  \wedge  ((\F  \text{ack}1) \U  \text{ack}0) $ \\ 
			\addlinespace[0em]
			& $ 2 $ & $ \G ( (\text{msg}0 \rightarrow\F (\text{ack}0))  \wedge  (\text{ack}0 \rightarrow\F (\text{msg}1))  \wedge  (\text{msg}1 \rightarrow\F (\text{ack}1)) ) $ & $ \G  ((\text{msg}0 \vee  (\G  (\text{msg}1 \vee  \text{ack}0)))  \rightarrow (\X  \text{ack}0)) $ \\ 
			\addlinespace[0em]
			& $ 3 $ & $ \G (\text{ack}1) $ & $ \G  \text{ack}1 $ \\ 
			\addlinespace[0em]
			& $ 4 $ & $ \bot $ & $ \bot $ \\ 
			\addlinespace[0em]
			& $ 5 $ & $ \top $ & $ \G  ((\text{ack}0 \vee  (\G  (\text{msg}0 \vee  \text{msg}1)))  \rightarrow (\X  \text{msg}1)) $ \\ 
			\addlinespace[0em]
			& $ 6 $ & $  \neg (\F (\text{msg}0 \U  \text{ack}0)  \wedge  \F (\text{ack}0 \U  \text{msg}1)  \wedge  \F (\text{msg}1 \U  \text{ack}1)) $ & $  \neg  (\text{ack}1 \vee  (\text{ack}0 \vee  (\text{msg}1 \vee  (\G  \text{msg}0)))) $ \\ 
			\addlinespace[0em]
			& $ 7 $ & $  \neg \F (\text{msg}0) $ & $ (\text{msg}1 \vee  \text{ack}0) \U  (\G  ( \neg  (\text{msg}0 \vee  (\text{msg}1 \vee  \text{ack}0)))) $ \\ 
			\addlinespace[0em]
			& $ 8 $ & $  \neg \F (\text{msg}1) $ & $ (\F  \text{ack}0) \U  (\G  ( \neg  (\text{ack}0 \vee  (\text{msg}0 \vee  \text{msg}1)))) $ \\ 
			\addlinespace[0em]
			& $ 9 $ & $  \neg \F (\text{ack}0) $ & $ (\F  (\text{msg}1 \vee  (\text{msg}0 \vee  \text{ack}0)))  \rightarrow (\text{msg}0  \wedge  ((\text{msg}1 \vee  (\text{msg}0 \vee  \text{ack}0)) \U  \text{msg}1)) $ \\ 
			\addlinespace[0em]
			& $ 10 $ & $  \neg \F (\text{ack}1) $ & $  \neg  (\text{msg}0 \vee  (\text{msg}1 \vee  (\text{ack}0 \vee  \text{ack}1))) $ \\ 
			\addlinespace[0em]
			
			\bottomrule
		\end{tabular}
	\end{center} 
\end{table}

We have considered a query $ \G(\text{ack}1) $ where  the learned explanation is also $ \G(\text{ack}1) $: if a word only contains ack$ 1 $ in every position, then according to Figure~\ref{fig:alternating-bit-protocol}, it is an accepting word and the RNN successfully learns it. We have considered another query $ \neg \F(\text{msg}0) $: the word does not contain any msg$ 0 $. {\framework} learns a PAC explanation $ (\text{msg}1 \vee  \text{ack}0) \U  (\G  ( \neg  (\text{msg}0 \vee  (\text{msg}1 \vee  \text{ack}0))))   $ for this query. The interpretation of this LTL formula is: if the word does not have msg$ 0 $ in any position then the RNN accepts a word where either msg$ 1 $ or ack$ 0 $ appears until in every position onwards ack$ 1 $ always appears. As shown in the automata of the ground truth in Figure~\ref{fig:alternating-bit-protocol}, if msg$ 0 $ is always absent in the word, then all accepting words should contain only ack$ 1 $ and the learned explanation of {\framework} has approximately captured this. We have similarly considered a query $ \neg \F(\text{ack}1) $ that accepts a word without ack$ 1 $. The learned (and expected) explanation of this query is false which is equivalent to the LTL formula $   \neg  (\text{msg}0 \vee  (\text{msg}1 \vee  (\text{ack}0 \vee  \text{ack}1))) $.  We next discuss the average performance of LTL explanations in comparison with DFAs in the following.

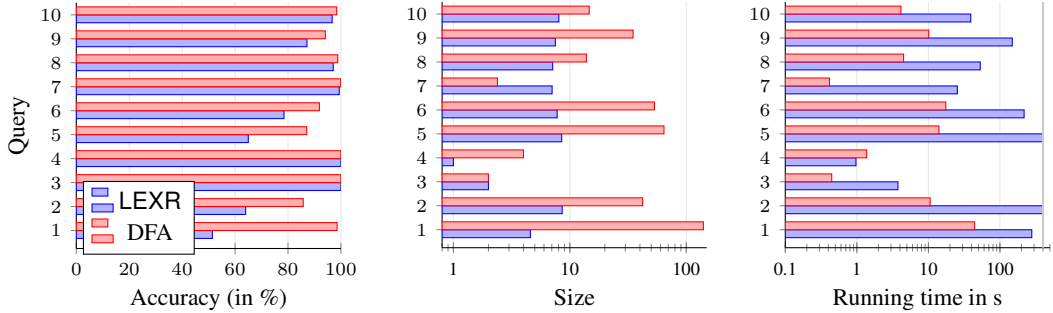
\begin{figure}[h]
	\centering

	\begin{tikzpicture}
	\begin{axis}[barplot2, width=51mm, height=48.5mm, xmin=0, ymin=0.25, ymax=10.5, xlabel={Accuracy (in \%)}, ylabel={Query}, legend columns=1, legend style={at={(.025, .275)}, anchor=north west, font=\footnotesize}]
	\addplot table [x=accuracy-LTL, y=query-no, col sep=comma] {csv/abp.csv};
	\addlegendentry{\framework}
	\addplot table [x=accuracy-DFA, y=query-no, col sep=comma] {csv/abp.csv};
	\addlegendentry{DFA}
	\end{axis}
	\end{tikzpicture}
	\hfill
	\begin{tikzpicture}
	\begin{axis}[barplot2, width=51mm, height=48.5mm, xmin=0.8, xmax=150, ymin = 0.25, ymax=10.5, xtick distance=10, xlabel={Size},xmode=log, log origin=infty,xtickten={-1,0, 1, 2}, xticklabels={0, 1, 10, 100}]
	\addplot table [x=size-LTL, y=query-no, col sep=comma] {csv/abp.csv};
	\addplot table [x=size-DFA, y=query-no, col sep=comma] {csv/abp.csv};
	\end{axis}
	\end{tikzpicture}
	\hfill
	\begin{tikzpicture}
	\begin{axis}[barplot2, width=51mm, height=48.5mm, xmin=.1, xmax=500, ymin = 0.25, ymax=10.5, xmode=log, log origin=infty, xlabel={Running time in s}, xtickten={-1,0, 1, 2}, xticklabels={0.1, 1, 10, 100}]
	\addplot table [x=time-LTL, y=query-no, col sep=comma] {csv/abp.csv};
	\addplot table [x=time-DFA, y=query-no, col sep=comma] {csv/abp.csv};
	\draw[color=gray!40, thick] (axis cs:400, .25) -- (axis cs:400, 12.75);
	\end{axis}
	\end{tikzpicture}
	\caption{Experimental results for eleven queries to a RNN for alternating bit protocol problem. Explanation size and running time are in the log scale.} 
	
	\label{fig:results_abp}
\end{figure}

In the left plot of Figure~\ref{fig:results_abp}, we show the average accuracy of both explanations. We have observed that LTL explanations have on average similar accuracy to that of DFAs in five out of ten queries. In the rest five queries, DFAs are better in accuracy. However, the average size of DFAs are higher than LTL explanations in eight out of ten queries as shown in the middle plot. More precisely, the result  DFAs are much larger in queries where the accuracy is also high. 
Finally, in the right plot, the average computation time of LTL learner is comparatively higher than that of DFA learner in most of the queries.

\bibliographystyle{splncs04}
\bibliography{draft}

\end{document}